\documentclass{svproc}
\usepackage{url}
\usepackage{graphicx}
\usepackage{textcomp}

\usepackage{hyperref}
\usepackage[utf8x]{inputenc}

\begin{document}

\mainmatter              
\title{Wind Power Projection using Weather Forecasts by Novel Deep Neural Networks}
\titlerunning{Deep Neural Network for Power Prediction}  
%
\makeatletter
\newcommand{\printfnsymbol}[1]{%
  \textsuperscript{\@fnsymbol{#1}}%
}

\author{Alagappan Swaminathan\inst{1}\thanks{Equal Contribution}, Venkatakrishnan Sutharsan\inst{1} \printfnsymbol{1}
\and Tamilselvi Selvaraj\inst{2}}
\authorrunning{Alagappan et al.} 
%
\tocauthor{Alagappan Swaminathan, Venkatakrishnan Sutharsan}
\institute{SSN College of Engineering, Kalavakkam, Chennai \\\
\email{\{alagappan16011, venkatakrishnansutharsan16119\}@eee.ssn.edu.in},\\ 
\and
Associate Professor, SSN College of Engineering, Kalavakkam, Chennai \email{tamilselvis@ssn.edu.in}}

\maketitle             

\begin{abstract}
The transition from conventional methods of energy production to renewable energy production necessitates better prediction models of the upcoming supply of renewable energy. In wind power production, error in forecasting production is impossible to negate owing to the intermittence of wind. For successful power grid integration, it is crucial to understand the uncertainties that arise in predicting wind power production and use this information to build an accurate and reliable forecast. This can be achieved by observing the fluctuations in wind power production with changes in different parameters such as wind speed, temperature, and wind direction, and deriving functional dependencies for the same. Using optimized machine learning algorithms, it is possible to find obscured patterns in the observations and obtain meaningful data, which can then be used to accurately predict wind power requirements . Utilizing the required data provided by the Gamesa's wind farm at Bableshwar, the paper explores the use of both parametric and the non-parametric models for calculating wind power prediction using power curves. The obtained results are subject to comparison to better understand the accuracy of the utilized models and to determine the most suitable model for predicting wind power production based on the given data set.
\end{abstract}


\section{Introduction}
Wind energy is undoubtedly one of the most fundamental pillars that supports the growth of mankind. It is the sustenance of modern economies and the key for global progress. For generations, the demand for wind energy has been met by conventional sources of energy: oil, coal, and petroleum.The widespread heavy usage of these non-renewable sources is accompanied by several concerns. Firstly, the most pressing consequence of the burning of fossil fuels is the growing climate crisis. The extraction as well as combustion of conventional energy sources results in the emission of greenhouse gases and pollutants which, in turn, impacts the air quality, contributes to global warming, and causes damage to the surrounding environment. In addition, the heavy reliance of global economies on fossil fuels for energy is cause for worry; since these sources are non-renewable, increased usage will eventually lead to their depletion, leaving the energy security of developed and developing nations to hang in the balance and restricting their energy independence. These environmental and public health concerns are further exacerbated by a rapidly growing population, improvements in standards of living, and increased industrialization which will cause a massive increase in the energy demand.

\section{Literature Survey}
\subsection{Wind Power in India}
India is relatively a newcomer to the global wind power market, compared to Denmark and the United States of America (USA). However, the growth of wind power in India in recent times has been tremendous, ranking the Indian wind power market as the fourth largest in the world after China, USA, and Germany, and the second largest wind power market in Asia. The leader in wind power in India is Tamil Nadu, with a wind power capacity that is approximately 29\% of the national total. The largest wind power plant in India is the Muppandal wind farm, which has a total capacity of 1500 MW.

\subsection{Indian Electricity Grid Code Requirements}
In India, the regulatory body of the power sector is the Central Electricity Regulatory Commission (CERC). The CERC has established regulations for power system operations known as the IEGC.According to these requirements, every wind farm with aggregate generation capacity of 50 MW and above must perform wind energy forecast with an interval of 15 minutes for the next 24 hours. In order to ensure that the forecasting models are aligned to minimize the actual MW deviations, CERC defines the percentage error normalized to the rated capacity of the plant. Wind farm operators are allowed to submit a total of 8 revisions throughout the day, up to 3 hours before the actual schedule. In doing so, the forecast is improved; as more data becomes available, forecast errors are minimized, consequently reducing economic losses.





\subsection{Available Prediction Methods}
Numerous models have been identified for the crucial task of predicting wind power production, each having its own characteristics. The performance of each model is dependent upon the nature of the data it is trained with, which is mostly site-specific. Thus, the accuracy of prediction of wind power production in a particular region or for a particular wind farm relies on selecting the appropriate model to perform the task The performance of each model is dependent upon the nature of the data it is trained with, which is mostly site-specific. Thus, the accuracy of prediction of wind power production in a particular region or for a particular wind farm relies on selecting the appropriate model to perform the task.

The models which are currently available include: 
\begin{itemize}
    \item Persistence Method
    \item Numerical Weather Prediction Methods
    \item Statistical and Neural Network Methods
\end{itemize}

\subsection{Persistence Method}
The persistence method operates on the simple assumption that the wind speed or wind power at a certain time in the future will be the same as it was when the forecast was made. In other words, this model assumes a high correlation between the present and future wind values. The persistence method is not only the simplest and most economical, but also more accurate than other wind forecasting methods - but only in ultra-short-term forecasting. This method is used in electrical utility for ultra-short-term forecasts. But its accuracy degrades rapidly as the time-scale of forecasting increases.\cite{Wu}

\subsection{Numerical Weather Prediction Methods}
This is a classical forecasting technique. Numerical Weather Prediction (NWP) models operate by solving conservation equations (mass, momentum, heat, water, etc.) numerically at given locations on a spatial grid. Future behaviour of the atmosphere is predicted using current conditions and past trends. This technique is based on the topology and layout of the wind farms, specifications of the wind turbines, and numerical weather predictions updated by meteorological services along with data from a large number of sources.\cite{Lange}

When attempting the task of prediction using a NWP model, the most critical step is the selection of the particular NWP model. There are several criteria which must be taken into consideration when making this choice: the geographical area, the resolution (both spatial and temporal), the forecast horizon, the accuracy required, the computation time, and the number of runs. The focus of research on prediction methods, until recently, has been NWP models. This is mainly because NWP models are dominant for long-term forecasting (several hours to days ahead).

However, it is important to note that these models are incredibly complex and take several hours to obtain a solution on a super computer. In addition, NWP models are also susceptible to inflicting uncertainties which can affect the accuracy of prediction. These uncertainties arise from model assumptions of initial conditions of the atmosphere, adopted parameterization methods, and post-processing techniques which do not always replicate the real-time conditions. Additionally, these assumptions inherently generalize the state of the atmosphere, which often ignores the local phenomenon and weather variability. These uncertainties eventually influence the performance of the prediction model at a regional scale. For short-term prediction models, therefore, NWP models are not preferred\cite{Basu}.

\subsection{Statistical and Neural Network Methods}
This method relies on learning from historical data. A vast amount of data is analyzed and meteorological processes are not explicitly represented. Instead, a link between the power produced in the past and the weather during those time periods is determined. This is then utilized for predicting the future output. The model is generally represented as a time series model or a dynamic model whose parameters are estimated by minimizing a cost function over historical training data set.

There are several techniques which are classified as statistical methods of prediction. Some of the most common are  the Box-Jenkins methodology, autoregressive (AR), moving average (MA), the autoregressive integrated moving average model (ARIMA), autoregressive moving average model (ARMA), and the use of the Kalman filter. These methods model the statistical relationship among the historical data using classical time series analysis. Soft computing and machine learning approaches such as  regression algorithms, , ANN, and support vector machine (SVM), to name a few, also come under this category. In particular, Foley et al. describe the approach of using ANNs as a data-driven approach\cite{Foley}. Before arriving at a selection of models to be studied in this report, the usage of prediction techniques in other studies was investigated.

Torres et al. obtained a 20\% error reduction compared to persistence to forecast average hourly wind speed for a 10 hour forecast at a number of locations using nine years of historical dating using an ARMA model\cite{Torres}. Using ARIMA models, Kavasseri and Seetharaman took significantly longer prediction horizons into account and predicted wind speeds of 1-day-ahead and 2-day-ahead periods, respectively. Through this method, the forecasting accuracy was improved by an average of 42\% compared to the persistence method. In doing so, they highlighted the necessity of scheduling dispatchable generation and tariffs in the day-ahead electricity market\cite{Kavasseri}. Additionally, Lei and Ran presented a hybrid model based on wavelet-decomposition and ARMA for short-term wind speed prediction for a wind farm\cite{Lei}.

In Brazil, Lira et al. relied on linear regression to predict wind speeds at three different altitudes in two different geographical areas, Paracuru and Camocim, describing spatial relationships of wind speed data between these two sites. The results of this approach demonstrated the strength of simple regression methods\cite{Lira}. On the other hand, a study performed by Kafazi et al. found polynomial curve fitting model to be a better choice for forecasting energy production applications\cite{Kafazi}.

Techniques such as ANNs and Neuro-fuzzy networks have been used increasingly in recent times. In a comprehensive comparison study on the application of adaptive linear element, back propagation, and radial basis function type neural networks, Li and Shi applied ANNs in 1-hour-ahead wind speed forecasting\cite{Li}. In a comparative analysis of three types of neural networks, namely multi-layer perceptron (MLP), simultaneous recurrent neural network (SRN), and Elman recurrent neural network, which were trained using particle swarm optimization(PSO), Welch et al. performed short-term prediction of wind speed\cite{Welch}. Additionally, Jursa and Rohrig presented an approach combining ANN and nearest-neighbor approaches in an optimization model\cite{Jursa}.

Statistical methods work best for short-term models, the reason being that forecasting errors could get accumulated quickly with increasing prediction horizons. It is also important to note that these models are not portable; they require the assistance of a domain expert to create an individualized model for each wind farm. Since there is a lack of portability, changes in weather conditions would warrant significant changes in the model itself, which in turn would require the attention of an expert for re-tuning the parameters of the statistical model\cite{Negnevitsky}.

\section{Novelty}
The Novel approach taken in this paper is to use other parameters which are wind direction and temperature to assist wind speed in predicting the output power. The major improvements are shown in R-Squared value which is a key parameter for prediction analysis which is basically the goodness of fit measure for learning models. The average R-Squared value of our predicted Linear Regression model is around 0.8708\% which is very high compared to R-Squared value of 0.692\%\cite{Kafazi}. Also, the Polynomial Regression of 5th order prove to be a better model with maximum R-Squared value of 0.96435\%  compared to 0.9342\% using  2nd degree Polynomial Regression by Kafazi et al\cite{Kafazi}. The Artificial Neural Network has shown an average R-Squared Value of 0.966075\% which is the best in this comparison. The Artificial Neural Network with Wind speed and direction as input has shown better learning compared to ANN with wind speed, direction and temperature as input indicating the minimal effect of temperature of wind in Wind Power Forecasting. Thus this novel approach has been proved to be better in performance compared to the other existing methods.

\section{Experimentation}
\subsection{Wind Power Prediction using Power Curve}
\begin{figure}
\centering
\includegraphics[width=250pt]{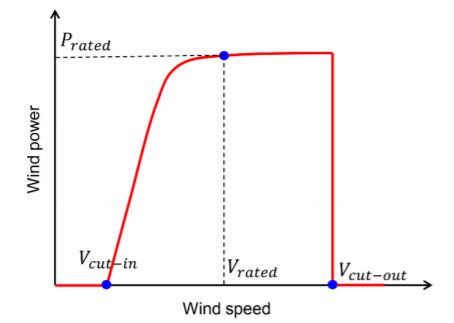}
\caption{Typical Wind Power Curve} \label{fig1}
\end{figure}

Before predicting wind power outputs, it is necessary to analyze power curves as they represent characters of wind turbine outputs. A power curve represents the relationship between the wind speed and the output power of a turbine. For a particular turbine, the power curve is provided by the turbine manufacturer. It is used for energy assessment and for monitoring the performance of the turbine. The Fig.~\ref{fig1} represents a typical power curve of a wind turbine.

\begin{figure}
\centering
\includegraphics[width=250pt]{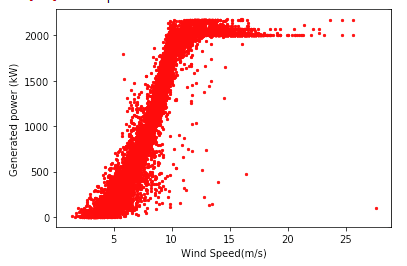}
\caption{Actual Representation of the Wind Power Curve} \label{fig2}
\end{figure}

Also we know the relation between the power generated and the wind speed to be known by
\begin{equation}
    P = 0.5\rho C_pAV^3
    \label{eqn1}
\end{equation}
where $\rho$ is density of the air, $C_p$ is the power co-efficient, A is the area wind turbine and V is the velocity of the wind. The Fig.~\ref{fig2} represents a actual power curve of a wind turbine.

Prediction models that are based on this equation suffer from a number of limitations. Firstly, the fraction of wind power that gets converted into electric power depends on a number of factors. Moreover, the variation of the factors in the equation with changing weather conditions, time, and turbine design are not taken into consideration; they are taken as constants. Finally, these models are cumbersome owing to the mathematical calculations, and do not provide accurate results\cite{Wadhvani}.

According to the law of conservation of mass and energy, not more than 59\% percent of the kinetic energy of wind can be captured. As explored in previous sections, the magnitude of wind power produced is dependent upon several factors such as wind speed, wind direction, air density, and turbine parameters.So, it is difficult to predict output power when creating a model that is based on equation~\ref{eqn1}.

Power curves provide a convenient method to perform the prediction task. But, since the manufacturer’s power curve is created under standard conditions, it may not be suitable for representing the realistic conditions of the particular site being considered for wind power prediction. Instead, it is possible to model power curves by deriving them using the actual data of wind speed and output power of the turbine, which makes the prediction more accurate\cite{Kusiak}.

Models can be classified as Parametric and Non-Parametric Models on the basis of the power curve.

\subsection{Parametric and Non-Parametric Models}

In parametric models, the relationship between the input and output, or the dependent and independent variable, is known. They are built from a set of mathematical equations that include parameters that must be adapted through a set of continuous data\cite{Pelletier}. The parameter values may be unknown, and are estimated from the training set. With developments in soft computing techniques, newer modelling methods have emerged known as learning methods. These are classified as non-parametric methods. Here, instead of assuming a physical or analytical relationship between the input and output data, a correlation is established based only on the data provided.An example of this would be ANNs, the use of which in predicting wind power production has been explored in this paper. In the extensive literature surveyed, none of the authors ever modelled a power curve with more than three inputs simultaneously.

\subsection{Linear Regression}
Linear regression is one of the most fundamental algorithms used for predictive analysis. It is utilized for determining a relationship between one dependent variable and one or more independent variables. By using this algorithm, we aim to obtain the best fit line or regression equation that can be used to make accurate predictions. One of the main advantages of using linear regression is its simplicity. This algorithm can be used to determine the strength of the factors being considered for prediction, conduct predictive analysis, and forecast trends. Yang et al. have proposed a linear regression based model for power prediction in which output power increases linearly.

As the name suggests, linear regression assumes a linear relationship between the input variables, say $x_i$, and the output variable, say y. This algorithm suggests that y can be calculated from a linear combination of input variables. In the linear equation formed, each input value is assigned a scale factor or coefficient. Additionally, another coefficient $\beta_0$ is included in the equation to provide an additional degree of freedom. In a simple regression problem, i.e., with a single input and output, the linear regression model would be expressed by the following equation:

\begin{equation}
    y = \beta_0 + \beta_1*x1 +\epsilon
    \label{eqn2}
\end{equation}

where y is the output value, $\beta_0$ the bias term,  $\beta_1$ is the slope coefficient, $x_i$ is the independent variable, and $\epsilon$ is a random error term.

The above representation simplifies the process of predicting output values for a given set of input values. For training the equation from data, there are several techniques which can be employed. Here, the Ordinary Least Squares method has been used for optimization and to minimize error. There is a problem of over-fitting which is predominantly due to increase in number of features giving rise to higher order polynomial equation.

\subsection{Polynomial Regression}

With linear regression, we attempt to find a linear relationship between the dependent variable and one or more independent variables. But, in many cases, the distribution of the data set is far more complex, and this linear model cannot be utilized efficiently to fit the data; doing so with a low value of error is incredibly difficult. Therefore, in these cases, it becomes a necessity to increase the complexity of the model. Polynomial regression is a predictive modelling technique wherein we attempt to fit a polynomial line in accordance with the non-linear relationship between the output variable and one or more input variables. It has been extensively used in literature to estimate the power curve of wind turbines. Shokrzadeh et al. proposed a polynomial regression based model for power prediction, where the relationship between the dependent and independent variables is curvilinear. The general equation of a polynomial regression model is:

\begin{equation}
    y = \beta_0+\beta_1*x_1+\beta_2*x_2+......+\beta_m*x_m + \epsilon
    \label{eqn3}
\end{equation}

One of the advantages of using this algorithm is that a broad range of functions can be fit under it. However, it is sensitive to outliers; the presence of even one or two outliers in the data set can impact the results determined from the non-linear analysis. 

\subsection{Artificial Neural Network}
\begin{figure}[]
\centering
\includegraphics[width=250pt]{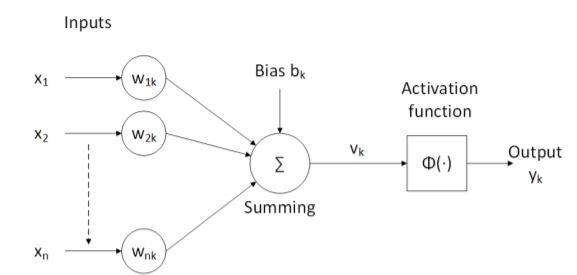}
\caption{Neuron of an ANN} \label{fig3}
\end{figure}

The working of ANN is inspired by biological nervous systems. ANNs are designed to imitate the working of the human brain. This is to determine the non-linear relationships between the input and the output data sets. Similar to the human brain, the fundamental building block of an ANN is the neuron, the structure of which is shown in Fig.~\ref{fig3}. Each neuron has an input signal $X_j$ which it alters and sends forward. Each neuron is interconnected to another via weighted synapses which decide the strength of the connection to the next step. The next step is the summing junction which adds the weighted input signals of the neuron. The output of the neuron is then limited using an activation function $\Phi$ which limits the output to a set finite range and returns an output. This can be expressed as:

\begin{equation}
    Y_k=\phi(u_k+b_k)
    \label{eqn4}
\end{equation}

\begin{figure}
\centering
\includegraphics[width=250pt]{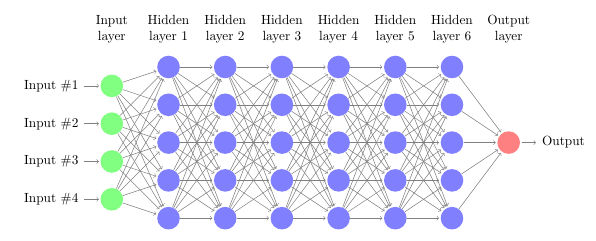}
\caption{General structure of an ANN with multiple hidden layers} \label{fig4}
\end{figure}

The proposed ANN model in this report is a  feed-forward neural network with an input layer, an output layer, and four hidden layers. A sample structure of an ANN with multiple hidden layers is shown in Fig.~\ref{fig4}. Further, two kinds of activation functions were utilized: Sigmoid and ReLU. Sigmoid functions allow for a smooth gradient, normalizing the output of each neuron in the ANN. Using the ReLU activation function makes the network more computationally efficient, allowing the network to converge very quickly. These activation functions were suitable here as the data to be processed as well as the expected outcome were not negative values. In general, there are several advantages in using ANNs. The key advantage of an ANN is its ability to learn and model non-linear and complex relationships. This makes ANNs suitable for capturing the non-linearity of the wind speed characteristic. After training, ANNs are capable of inferring unseen relationships, thus making the model generalize and predict on unseen data.

Additional advantages of an ANN are its ability to work with incomplete knowledge and its fault tolerance. After training the ANN, the network may produce an output even if some of the information is incomplete; the loss of performance depends on the
importance of the missing information. However, a disadvantage of using an ANN is its black-box approach; it is difficult to understand the exact internal working of the ANN, rendering its behaviour unexplained. This reduces trust in the network. In addition, there is no specific rule for determining the structure of an ANN; the appropriate structure is determined through trial and error.

\section{Evaluation of Models}
It is difficult to accurately measure the performance of different prediction models merely by comparing their output plots. Hence, in this report, the performance of the proposed models have been quantified using the following evaluation metrics:

\begin{itemize}
    \item Mean Absolute Error (MAE)
    \item Root Mean Square Error(RMSE)
    \item R Squared Value
\end{itemize}

\subsection{Mean Absolute Error}
The MAE is defined as the average of the difference between the predicted and actual values in the test - in other words, the average prediction error. It is expressed as in equation~\ref{eqn5}.

\begin{equation}
    MAE=1/N*\sum\limits_{i=1}^N\mathopen|p_i - p_i^{predicted}\mathclose|
    \label{eqn5}
\end{equation}

Where $ p_i^{predicted}$ is the power output and $p_i$ is the actual power output at time i, respectively, and N is the number of forecast samples.

\subsection{Root Mean Square Error}
The RMSE is defined as the square root of the mean squared error between the predicted value and the actual value. It is expressed as in equation~\ref{eqn6}.

\begin{equation}
    RMSE=\sqrt{(1/N*\sum\limits_{i=1}^N\mathopen|(p_i - p_i^{predicted})^2\mathclose|}
    \label{eqn6}
\end{equation}

Where $ p_i^{predicted}$ is the power output and $p_i$ is the actual power output at time i, respectively, and N is the number of forecast samples.

\subsection{R Squared Value}
Correlation, or ’R’, is a number between 1 and -1. A value of +1 indicates that an increase in an input value x results in an increase in an output value y, -1 indicates that an increase in x results in a decrease in y, and 0 means that there is no relationship between x and y. R square, like correlation, shows us how related two entities are, but is easier to interpret than correlation. R squared can be defined as percentage of variation in the dependent and independent variables. It is represented as in equation ~\ref{eqn7}.

\begin{equation}
    R^2=1-(\sum\limits_{i=1}^N(p_i - p_i^{predicted})^2)/(\sum\limits_{i=1}^N(p_i - p_mean)^2)
    \label{eqn7}
\end{equation}

Where $ p_i^{predicted}$ is the power output and $p_i$ is the actual power output at time i, respectively, and N is the number of forecast samples. R squared values are between 0 and 1, or 0\% and 100\%. The higher the value, the more fitting the prediction model.

\section{Results and Discussion}
\subsection{The Dataset}
The necessary data was provided by Gamesa's wind farm at Bableshwar. This institution, formerly known as the Centre for Wind Energy Technology (C-WET) and located in Chennai, Tamil Nadu, serves as a focal point in the improvement and development of the entire spectrum of the wind energy sector in India. The given data represents the parameters of wind power production at a particular substation. It spans over the period from January to December of 2019 and records wind speed, wind power output, temperature, and wind direction at 15-minute intervals. There were a total of 30090 entries for each parameter. It is also interesting to note that at the substation from which the data was collected, August was a particularly windy month, likely due to the monsoons, and the power generated during this month was higher than the previous months.

\subsection{Impact of Wind speed, Wind direction and Temperature on Wind power production}
\begin{figure}
\centering
\includegraphics[width=250pt]{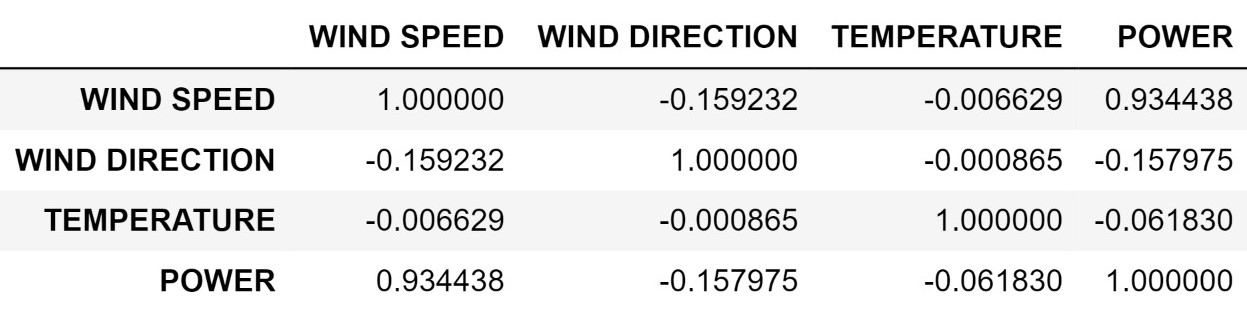}
\caption{Correlation of the effects of wind speed, wind direction, and temperature on wind power production} \label{fig5}
\end{figure}

For the prediction of wind power production, three factors have been taken into consideration: wind speed, wind direction, and temperature. Before using the respective data to train the proposed models, the extent of the influence of these parameters on wind power production was first explored. Fig.~\ref{fig5} presents a matrix-like correlation, relaying the extent to which wind speed, wind direction, and temperature are correlated to each other and the generated wind power, based on the data provided by Gamesa's wind farm at Bableshwar. The greater the value at the intersection of two parameters, the more inter-related are the two parameters; the maximum value of correlation is 1. As explored in previous sections, wind power production is highly dependent upon wind speed. This relationship is reflected in Fig.~\ref{fig5}, with a value of 0.934438 at their intersection. We can also see that wind direction and temperature have a negligible impact on wind power production as the corresponding values in the figure are incredibly low. This information provided by Fig.~\ref{fig5} can also be represented visually by means of a heat map.

\begin{figure}
\centering
\includegraphics[width=250pt]{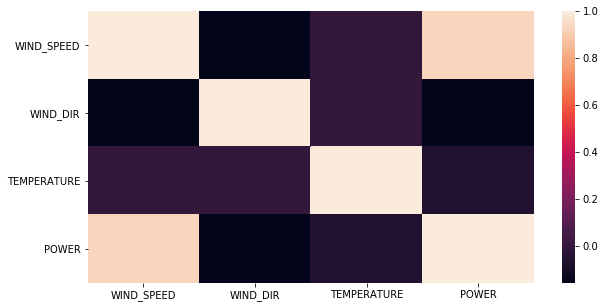}
\caption{Heatmap to visually represent the correlation matrix} \label{fig6}
\end{figure}

Fig.~\ref{fig6}, which has been generated using Python, represents the correlation between any two of the considered parameters using the colour of the tile at the intersection of two parameters. As per the colour scale provided in the figure, a lighter colour represents greater correlation and a darker colour represents negligible correlation. Here, too, we see the great influence of wind speed upon wind power production and the negligible effect of wind direction and temperature on wind power production. Nevertheless, wind direction and temperature have been taken into consideration for training the proposed models in order to analyze the impact of these parameters in reducing the error of prediction.

\subsection{Linear Regression}
It establishes a linear relationship between the dependent and independent variables.First, the necessary Python libraries were imported. Then, the given data set was loaded. The data set contains all the values in the form of a table. Variables X and y were trained with data from the relevant columns of the data table. By setting the value of ’test size’, we were able to vary the proportion of the data set being used for training and testing the model.The ’random state’ parameter was used to randomize the splitting of the data into train and test indices. Following this, a linear regression object was created. The corresponding MAE, RMSE, and R squared values were then calculated. 

\begin{figure}
\centering
\includegraphics[width=250pt]{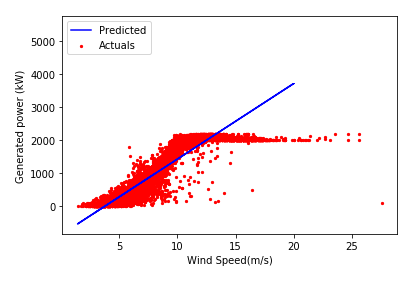}
\caption{Power curve fitting when the proposed linear regression model is trained using 85\% of the given data} \label{fig7}
\end{figure}

Fig.~\ref{fig7} was plotted using 85\% of the given data when considering wind speed, wind direction, and temperature.Given that this is a linear regression model, the expected relationship between wind power production and the influencing parameters was also linear. This is confirmed by the interpretation of Fig.~\ref{fig7}. It compares the power curve that corresponds to the predicted values of generated power and wind speed with the power curve that corresponds to the actual values of generated power and wind speed. The scattered red region represents the actual values plotted between wind speed and wind power, and the blue line represents the power curve for the predicted value. Clearly, the power curve corresponding to the predicted values does not fit well against the power curve corresponding to the actual values. From this figure, one can interpret that the accuracy of the proposed linear regression model is dismal for the given data set.

\begin{figure}
\centering
\includegraphics[width=250pt]{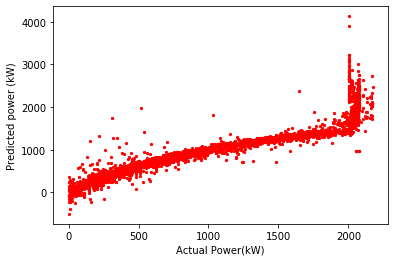}
\caption{Predicted power output vs. Actual power output using linear regression} \label{fig8}
\end{figure}
In Fig.~\ref{fig8}, the predicted power output has been plotted against the actual power output in order to gauge the accuracy of the predicted values. As can be observed, the resulting graph is not a straight linear line between the x and y axes. Hence, one can infer that the proposed linear regression model does not provide a sufficiently
accurate prediction based on the given data set.
\begin{table}
\centering
\caption{Evaluation of the proposed linear regression model when trained with wind speed data only}
\label{tab2}
{
\centering
\begin{tabular}{ |p{2cm}|p{2cm}|p{2cm}|p{2cm}|p{2cm}|}
 \hline
Training set size &Testing set size &MAE  &RMSE &R Squared\\
\hline
0.95 & 0.05 & 178.33192 & 253.35265 & 0.86980 \\
\hline
0.9 & 0.1 & 176.90861 & 247.66174 & 0.87571 \\
\hline
0.85 & 0.15 & 175.74426 & 247.39705 & 0.87627 \\
\hline
0.8 & 0.2 & 176.90189 & 262.27704 & 0.86103 \\
\hline
0.75 & 0.25 & 177.02434 & 259.97856 & 0.86345 \\
\hline
0.7 & 0.3 & 176.51930 & 257.89664 & 0.86533 \\
\hline
\end{tabular}
}
\end{table}

\begin{table}
\centering
\label{tab3}
\caption{Evaluation of the proposed linear regression model when trained with wind speed and wind direction data only}
\label{tab3}
{
\begin{tabular}{ |p{2cm}|p{2cm}|p{2cm}|p{2cm}|p{2cm}|}
 \hline
Training set size &Testing set size &MAE  &RMSE &R Squared\\
\hline
        0.95 & 0.05 & 178.81581
         & 253.11586 & 0.87107 \\
        \hline
         
        0.9 & 0.1 & 177.40251 & 246.68685 & 0.87668 \\
        \hline
        
        0.85 & 0.15 & 176.37347 & 246.52254 & 0.87714 \\
        \hline
        
        0.8 & 0.2 & 177.55531 & 261.27449 & 0.86209 \\
        \hline
        
        0.75 & 0.25 & 177.90095 & 258.99536 & 0.86448 \\
        \hline
        
        0.7 & 0.3 & 177.42095 & 256.85976 & 0.866411 \\
        \hline

\end{tabular}
}
\end{table}

\begin{table}
\centering
\caption{Evaluation of the proposed linear regression model when trained with wind speed and temperature data only}
\label{tab4}
{
\centering
\begin{tabular}{ |p{2cm}|p{2cm}|p{2cm}|p{2cm}|p{2cm}|}
\hline
Training set size &Testing set size &MAE  &RMSE &R Squared\\
\hline
        0.95 & 0.05 & 175.36880         & 250.02119 & 0.87320 \\
        \hline
         
        0.9 & 0.1 & 173.18357 & 243.63749 & 0.87971 \\
        \hline
        
        0.85 & 0.15 & 172.45955 & 243.95124 & 0.87969 \\
        \hline
        
        0.8 & 0.2 & 173.64332 & 258.94650 & 0.86453 \\
        \hline
        
        0.75 & 0.25 & 173.98111 & 256.71224 & 0.86686 \\
        \hline
        
        0.7 & 0.3 & 173.52368 & 254.57559 & 0.86877 \\
        \hline

\end{tabular}
}
\end{table}

\begin{table}
\centering
\caption{Evaluation of the proposed linear regression model when trained with wind speed, wind direction, and temperature data}
\label{tab17}
{
\centering
\begin{tabular}{ |p{2cm}|p{2cm}|p{2cm}|p{2cm}|p{2cm}|}
 \hline
Training set size &Testing set size &MAE  &RMSE &R Squared\\
\hline
0.95 &0.05 &175.84133 &249.02472 &0.87421\\
\hline
0.90 &0.10 &173.62791 &242.88975 &0.88045\\
\hline
0.85 &0.15 &173.08321 &243.29391 &0.88034\\
\hline
0.80 &0.20 &174.29879 &258.14871 &0.86537\\
\hline
0.75 &0.25 &174.82482 &255.90645 &0.86769\\
\hline
0.70 &0.30 &174.40050 &253.70599 &0.86967\\
\hline

\end{tabular}
}
\end{table}

The first column of each table namely TABLE~\ref{tab2}, TABLE~\ref{tab3}, TABLE~\ref{tab4}, TABLE~\ref{tab17} represents the fraction of the data set which was used to train the model. The values in this column vary from 0.95 to 0.7 in decrements of 0.05. The second column represents the fraction of the data set which was used for validating the values predicted by the model. The last three columns represent the corresponding MAE, RMSE, and R squared values which were considered for evaluating the performance of the model.The evaluation metrics confirm that the accuracy of the proposed linear regression model is not impressive; while the R squared value is above 0.8 under all conditions, a more reliable prediction model is certainly preferable. Further, if one observes the variation in the values of the evaluation columns, one can notice that the best values, i.e., least error, is obtained when the training set size is 0.85.So, for the particular sub-station taken into consideration and for the corresponding data set, the proposed linear regression model works best with a training set size ranging from 0.85 to 0.9 when trained with wind speed, wind
direction, and temperature data.

\subsection{Polynomial Regression}
With the proposed polynomial regression model, the expectation is to obtain a power curve from the predicted values which fits closer to the actual power curve, unlike the proposed linear regression model. When using polynomial regression, one accounts for the complexity of the distribution of a particular data set. So, with the proposed polynomial regression model, the non-linear relationship between dependent and independent variables is accounted for. The data set contains all the values in the form of a table. Variables X and y were trained with data from the relevant columns of the data table. By setting the value of ’test size’, we were able to vary the proportion of the data set being used for training and testing the model. The ’random state’ parameter was used to randomize the splitting of the data into train and test indices. Then, a polynomial features object was created, followed by a linear regression object. Since the degree of the polynomial also influences the accuracy, the degree of the polynomial was varied. 

\begin{figure}
\centering
\includegraphics[width=250pt]{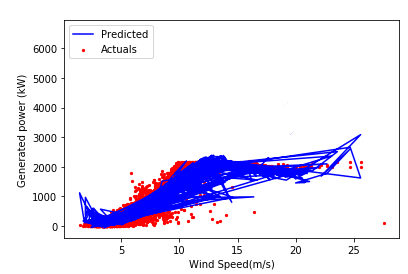}
\caption{Power curve fitting when the proposed polynomial regression model is trained
using 85\% of the given data and the degree is 5} \label{fig9}
\end{figure}

This value can vary from 2 to (n-1), where n is the number of data points. The corresponding MAE, RMSE, and R squared values were then calculated and printed. Fig.~\ref{fig9} was then plotted,using 85\% of the given data when considering wind speed, temperature, and wind direction. Since this is a polynomial regression model, the expected relationship between wind power production and the influencing parameters is non-linear. This is reflected in Fig.~\ref{fig9} which compares the power curve corresponding to the predicted values - a sigmoid curve - with that of the actual values. Clearly, one can observe that the power curve corresponding to the predicted values fits much better with the actual power curve than the proposed linear regression model. From this, one can deduce that the accuracy of prediction of the proposed polynomial regression model is much greater than that of the proposed linear regression model.

\begin{figure}
\centering
\includegraphics[width=250pt]{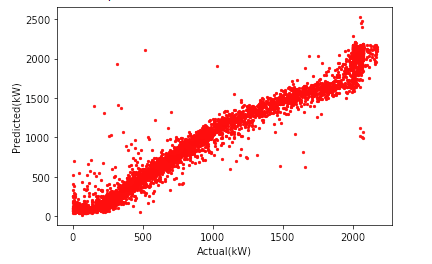}\
\caption{Predicted power output vs. Actual power output using polynomial regression} \label{fig10}
\end{figure}

In Fig.~\ref{fig10}, the predicted power output has been plotted against the actual power output in order to gauge the accuracy of the predicted values. As can be observed, the obtained plot is mostly straight and linear. This is another indication that the proposed polynomial regression model has predicted values that are much more accurate than those of the proposed linear regression model.

\begin{table}
\centering
\caption{Evaluation of the proposed polynomial regression model when trained with wind speed data only}
\label{tab6}
    \begin{tabular}{ |p{2cm}|p{2cm}|p{2cm}|p{2cm}|p{2cm}|p{2cm}|}
          \hline
            \textbf{Train Set Size}&\textbf{Test Set Size} &\textbf{Degree}& \textbf{MAE}&\textbf{RMSE}&\textbf{R Squared} \\
        \hline 
         &  & 2 & 163.36605 & 218.15300 & 0.90346 \\ \cline{3-6}
         0.95 & 0.05 & 3 & 138.87893 & 196.80986 & 0.92143 \\ \cline{3-6}
         
         & & 4 & 100.38071 & 164.39565 & 0.94518 \\ \cline{3-6}
          
         & & 5 & 100.98805 & 163.82108 & 0.94556 \\
         \hline
         
         & & 2 & 159.26711 & 211.14781 & 0.90966 \\ \cline{3-6}
         
         0.9 & 0.1 & 3 & 136.72468 & 190.29981 & 0.92662 \\ \cline{3-6}
          
          & & 4 & 98.50828 & 155.15862 & 0.95122 \\ \cline{3-6}
          
          & & 5 & 98.10501 & 152.98460 & 0.954574 \\
          \hline
          
          & & 2 & 158.44036 & 207.23645 & 0.91318 \\ \cline{3-6}
          
          0.85 & 0.15 & 3 & 136.08039 & 190.7105 & 0.92647 \\ \cline{3-6}
          
          & & 4 & 97.07183 & 148.28485 & 0.95555 \\ \cline{3-6}
          
          & & 5 & 96.60666 & 146.14452 & 0.95682 \\
          \hline
          
          & & 2 & 160.11458 & 209.16813 & 0.91161 \\ \cline{3-6}
          
         0.8 & 0.2 & 3 & 133.72531 & 211.61354 & 0.90953 \\ \cline{3-6}
          
          & & 4 & 96.93708 & 184.77183 & 0.93103 \\ \cline{3-6}
          
          & & 5 & 96.35479 & 161.74937 & 0.94714 \\
          \hline 
          
          & & 2 & 160.24290 & 209.25208 & 0.91154 \\ \cline{3-6}
          
         0.75 & 0.25 & 3 & 133.50883 & 205.46278 & 0.91472 \\ \cline{3-6}
          
          & & 4 & 96.67450 & 178.46419 & 0.93566 \\ \cline{3-6}
          
          & & 5 & 96.35426 & 158.54910 & 0.94922 \\
          \hline
          
          & & 2 & 161.04756 & 209.03238 & 0.91153 \\ \cline{3-6}
          
         0.7 & 0.3 & 3 & 133.81947 & 200.42754 & 0.91866 \\ \cline{3-6}
          
          & & 4 & 95.82811 & 172.41005 & 0.93981 \\ \cline{3-6}
          
          & & 5 & 95.69629 & 155.00109 & 0.95135 \\        \hline

        \end{tabular}
\end{table}

\begin{table}
\centering
\caption{Evaluation of the proposed polynomial regression model when trained with wind speed and wind direction data only}\label{tab7}
    \begin{tabular}{ |p{2cm}|p{2cm}|p{2cm}|p{2cm}|p{2cm}|p{2cm}|}
          \hline
            \textbf{Train Set Size}&\textbf{Test Set Size} &\textbf{Degree}& \textbf{MAE}&\textbf{RMSE}&\textbf{R Squared} \\
        \hline 
        
       &  & 2 & 157.05884 & 209.08797 & 0.91132 \\ \cline{3-6}
         0.95 & 0.05 & 3 & 131.54443 & 186.36933 & 0.92954 \\ \cline{3-6}
         
         & & 4 & 97.45729 & 158.21575 & 0.94922 \\ \cline{3-6}
          
         & & 5 & 95.88115 & 151.38449 & 0.95351 \\
         \hline
         
         & & 2 & 152.54157 & 200.85121 & 0.91825 \\ \cline{3-6}
         
         0.9 & 0.1 & 3 & 128.39435 & 179.30980 & 0.93485 \\ \cline{3-6}
          
          & & 4 & 95.04405 & 151.07935 & 0.95375 \\ \cline{3-6}
          
          & & 5 & 92.51458 & 142.59749 & 0.95879 \\
          \hline
          
          & & 2 & 151.61996 & 196.97673 & 0.92156 \\ \cline{3-6}
          
          0.85 & 0.15 & 3 & 127.66517 & 179.62605 & 0.93477 \\ \cline{3-6}
          
          & & 4 & 91.97351 & 141.40261 & 0.95958 \\ \cline{3-6}
          
          & & 5 & 90.24005 & 135.03130 & 0.96314 \\
          \hline
          
          & & 2 & 149.56589 & 194.20941 & 0.92380 \\ \cline{3-6}
          
         0.8 & 0.2 & 3 & 124.87827 & 196.10105 & 0.92231 \\ \cline{3-6}
          
          & & 4 & 89.07193 & 172.24278 & 0.94006 \\ \cline{3-6}
          
          & & 5 & 88.52886 & 147.10882 & 0.95628 \\
          \hline 
          
          & & 2 & 152.86237 & 198.26405 & 0.92058 \\ \cline{3-6}
          
         0.75 & 0.25 & 3 & 124.56049 & 191.61831 & 0.92582 \\ \cline{3-6}
          
          & & 4 & 92.46087 & 170.92146 & 0.94098 \\ \cline{3-6}
          
          & & 5 & 90.32096 & 144.50929 & 0.95781 \\
          \hline
          
          & & 2 & 153.90863 & 198.8654 & 0.92007 \\ \cline{3-6}
          
         0.7 & 0.3 & 3 & 124.96455 & 187.43514 & 0.92886 \\ \cline{3-6}
          
          & & 4 & 93.75799 & 171.02757 & 0.94077 \\ \cline{3-6}
          
          & & 5 & 90.21331 & 143.04789 & 0.95857 \\        \hline
        \end{tabular}
\end{table}

\begin{table}
\centering
\caption{Evaluation of the proposed polynomial regression model when trained with wind speed and temperature data only}\label{tab8}
    \begin{tabular}{ |p{2cm}|p{2cm}|p{2cm}|p{2cm}|p{2cm}|p{2cm}|}
          \hline
            \textbf{Train Set Size}&\textbf{Test Set Size} &\textbf{Degree}& \textbf{MAE}&\textbf{RMSE}&\textbf{R Squared} \\
        \hline 
        
       &  & 2 & 161.40320 & 214.77899 & 0.90643 \\ \cline{3-6}
         0.95 & 0.05 & 3 & 136.60721 & 193.60344 & 0.92397 \\ \cline{3-6}
         
         & & 4 & 98.33635 & 161.14955 & 0.94732 \\ \cline{3-6}
          
         & & 5 & 98.18766 & 158.92774 & 0.94877 \\
         \hline
         
         & & 2 & 156.99932 & 207.33262 & 0.91289 \\ \cline{3-6}
         
         0.9 & 0.1 & 3 & 133.73842 & 186.66654 & 0.92939 \\ \cline{3-6}
          
          & & 4 & 96.32098 & 151.98026 & 0.95319 \\ \cline{3-6}
          
          & & 5 & 95.5189 & 148.08611 & 0.95556 \\
          \hline
          
          & & 2 & 156.26280 & 203.63869 & 0.91617 \\ \cline{3-6}
          
          0.85 & 0.15 & 3 & 133.73842 & 187.10393 & 0.92923 \\ \cline{3-6}
          
          & & 4 & 94.96499 & 144.78325 & 0.95762 \\ \cline{3-6}
          
          & & 5 & 94.20845 & 141.38741 & 0.95959 \\
          \hline
          
          & & 2 & 157.82921 & 205.42804 & 0.91474 \\ \cline{3-6}
          
         0.8 & 0.2 & 3 & 130.82381 & 207.38387 & 0.91311 \\ \cline{3-6}
          
          & & 4 & 94.86253 & 189.62197 & 0.92736 \\ \cline{3-6}
          
          & & 5 & 94.00498 & 181.79505 & 0.93323 \\
          \hline 
          
          & & 2 & 158.03800 & 205.65515 & 0.91455 \\ \cline{3-6}
          
         0.75 & 0.25 & 3 & 130.71028 & 201.46466 & 0.91800 \\ \cline{3-6}
          
          & & 4 & 94.74571 & 182.64485 & 0.93261 \\ \cline{3-6}
          
          & & 5 & 94.07443 & 175.65365 & 0.93767 \\
          \hline
          
          & & 2 & 158.79705 & 205.48684 & 0.91450 \\ \cline{3-6}
          
         0.7 & 0.3 & 3 & 131.07367 & 196.62502 & 0.92172 \\ \cline{3-6}
          
          & & 4 & 93.75628 & 175.81823 & 0.93741 \\ \cline{3-6}
          
          & & 5 & 93.17997 & 168.08954 & 0.94279 \\        \hline
        \end{tabular}
\end{table}

\begin{table}
\centering
\caption{Evaluation of the proposed polynomial regression model when trained with wind speed, wind direction, and temperature data}\label{tab9}
    \begin{tabular}{ |p{2cm}|p{2cm}|p{2cm}|p{2cm}|p{2cm}|p{2cm}|}
          \hline
            \textbf{Train Set Size}&\textbf{Test Set Size} &\textbf{Degree}& \textbf{MAE}&\textbf{RMSE}&\textbf{R Squared} \\
        \hline 
        
        &  & 2 & 157.11845 & 208.03268 & 0.91221 \\ \cline{3-6}
         0.95 & 0.05 & 3 & 131.45079 & 185.88187 & 0.92991 \\ \cline{3-6}
         
         & & 4 & 98.05413 & 158.74838 & 0.94888 \\ \cline{3-6}
          
         & & 5 & 93.73527 & 148.11939 & 0.95549 \\
         \hline
         
         & & 2 & 152.21444 & 199.90720 & 0.91902 \\ \cline{3-6}
         
         0.9 & 0.1 & 3 & 128.37896 & 178.86032 & 0.93517 \\ \cline{3-6}
          
          & & 4 & 95.80909 & 151.64924 & 0.9539 \\ \cline{3-6}
          
          & & 5 & 90.57723 & 139.46847 & 0.96058 \\
          \hline
          
          & & 2 & 151.24862 & 196.21854 & 0.92216 \\ \cline{3-6}
          
          0.85 & 0.15 & 3 & 127.38479 & 179.08433 & 0.93516 \\ \cline{3-6}
          
          & & 4 & 92.61943 & 141.62518 & 0.95945 \\ \cline{3-6}
          
          & & 5 & 88.83963 & 132.79841 & 0.96435 \\
          \hline
          
          & & 2 & 149.21291 & 193.62416 & 0.92426 \\ \cline{3-6}
          
         0.8 & 0.2 & 3 & 124.83034 & 195.19840 & 0.92302 \\ \cline{3-6}
          
          & & 4 & 89.04671 & 182.19318 & 0.93294 \\ \cline{3-6}
          
          & & 5 & 87.45523 & 166.16806 & 0.944217 \\
          \hline 
          
          & & 2 & 152.43461 & 197.55189 & 0.92116 \\ \cline{3-6}
          
         0.75 & 0.25 & 3 & 124.48798 & 190.55189 & 0.92642 \\ \cline{3-6}
          
          & & 4 & 93.74098 & 181.54354 & 0.93342 \\ \cline{3-6}
          
          & & 5 & 89.02249 & 160.86614 & 0.94772 \\
          \hline
          
          & & 2 & 153.48259 & 197.94288 & 0.92067 \\ \cline{3-6}
          
         0.7 & 0.3 & 3 & 124.91562 & 186.62300 & 0.92948 \\ \cline{3-6}
          
          & & 4 & 95.48558 & 181.98448 & 0.93294 \\ \cline{3-6}
          
          & & 5 & 88.21719 & 154.34707 & 0.95176 \\        \hline
        \end{tabular}
\end{table}

The first column of each table namely TABLE~\ref{tab6}, TABLE~\ref{tab7}, TABLE~\ref{tab8}, TABLE~\ref{tab9} represents the fraction of the given data which was used to train the model. The values in this column range from 0.95 to 0.7 in decrements of 0.05. The second column represents the fraction of the data set which was used for validating the values predicted by the model, i.e., the data remaining apart from the training data. For evaluating the proposed polynomial regression model, the degree of the polynomial was also considered an important parameter to study its influence on the accuracy of the values predicted by the model. This is what the third column represents. The degree of the polynomial regression model was varied from 2 to 5. This was done to see if an increase in complexity of the polynomial accounts for the complexity of the relationship between generated power and the three considered parameters and thus provides a more accurate result. The degree of the polynomial was not increased beyond 5 in order to avoid over-fitting the model and thus compromising the reliability of the model. The last three columns of the table explain the performance of the model in terms of MAE, RMSE, and R squared values respectively. For every training set size, the accuracy consistently increased with an increase in the degree of the polynomial. The MAE and RMSE values decreased and the R squared values increased with an increase in degree.

This indicates that lower degree polynomials were not ideal to represent the complexity of the relationship between generated wind power and the parameters taken into consideration, i.e., wind speed, wind direction, and temperature. So, one can observe that a fifth degree polynomial is best suited for accurate prediction using the proposed polynomial regression model. Further, under all four testing conditions, the best R squared values were obtained when the degree of the polynomial was 5 and the training set size was 85\% of the total data taken into consideration for each case. One can infer that a greater training set size would have contributed to an increase in error, and a lesser training set size would not have provided sufficient data for increased accuracy. The most accurate values predicted by the proposed polynomial regression value were obtained when the model was trained with wind speed, temperature data and wind direction, the training set size was 85\% of the given data, and the degree was 5; the R squared value under these conditions is 0.96435. This makes the proposed polynomial regression model much more suitable and reliable for the task of predicting wind power output for the sub-station taken under consideration than the previous model.

\subsection{Artificial Neural Network}
\begin{figure}
\centering
\includegraphics[width=250pt]{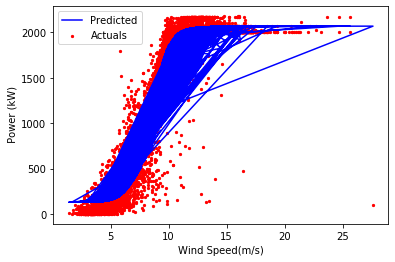}
\caption{Power curve fitting using the proposed ANN model when trained with 85\% of the given data} \label{fig11}
\end{figure}

The ANN is a non-parametric model of predicting wind power production.In this proposed model, rather than defining the relationship between wind power production and the considered parameters, the model is trained with the provided data and the relationship between the dependent and independent factors is derived by the model. This makes it much easier to establish a correlation based on the provided data.Variables X and y were trained with data from the relevant columns of the data table. By setting the value of ’test size’, we were able to vary the proportion of the data set being used for training and testing the model. The ’random state’ parameter was used to randomize the splitting of the data into train and test indices. Then, an ANN with four hidden layers using sigmoid and ReLu activation functions was created, compiled, trained with the given data, and tested. The corresponding MAE, RMSE, and R squared values were then calculated and printed. Fig.~\ref{fig11} was obtained using 85\% of the given data when considering wind speed, wind direction, and temperature.

Since ANNs have a black-box approach, making it difficult to understand the inner workings of the network and the relationship established between the input and output variables, the proposed feed-forward ANN model was created by trial-and-error methods. In the process, it was found that a four-layered ANN was optimal for the prediction task. Further, 20 epochs were found to be optimal for training the proposed ANN model. One can infer here that a lesser number of epochs would have resulted in under-fitting and more epochs would have caused over-fitting.

Fig.~\ref{fig11} provides a comparison between the power curve corresponding to the values predicted by the proposed ANN model and the actual power curve. Here, one can observe that the power curve corresponding to the predicted values is a sigmoid graph and fits very close to the actual power curve - even better than that obtained with the proposed polynomial regression model. From this, one can infer that the accuracy of prediction of the proposed ANN model is much greater than that of the proposed linear regression model and is likely to be greater than that of the proposed polynomial regression model.

\begin{figure}
\centering
\includegraphics[width=250pt]{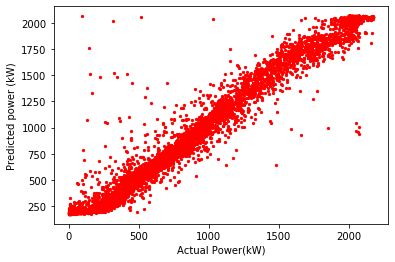}
\caption{Predicted power output vs. actual power output using ANN} \label{fig12}
\end{figure}

In Fig.~\ref{fig12}, one can see the predicted power output plotted against the actual power output. The graph is straight and linear - even more so than that obtained from the proposed polynomial regression model, which is another indicator that
the proposed ANN model is likely to have more accuracy than the polynomial regression model. This inference is substantiated by the following tables namely  TABLE~\ref{tab10}, TABLE~\ref{tab11}, TABLE~\ref{tab12}, TABLE~\ref{tab13}

Similar to the evaluation table corresponding to the proposed linear regression model, the table has five columns. The first column represents the fraction of the data set which was used to train the model. The values in this column vary from 0.95 to 0.7 in decrements of 0.05. The second column represents the fraction of the data set which was used for validating the values predicted by the model. The last three columns represent the corresponding MAE, RMSE, and R squared which were considered for evaluating the model.

\begin{table}
\centering
\caption{Evaluation of the proposed ANN model when trained with wind speed data only}\label{tab10}
{
\centering
\begin{tabular}{ |p{2cm}|p{2cm}|p{2cm}|p{2cm}|p{2cm}|}
\hline
Training set size &Testing set size &MAE  &RMSE &R Squared\\
\hline
 0.95 & 0.05 & 77.56814 & 145.24016 & 0.95721 \\
        \hline
         
        0.9 & 0.1 & 75.75352 & 134.63276 & 0.96327 \\
        \hline
        
        0.85 & 0.15 & 73.47237 & 125.15547 & 0.96833 \\
        \hline
        
        0.8 & 0.2 & 75.05013 & 127.97741 & 0.96691 \\
        \hline
        
        0.75 & 0.25 & 78.28338 & 124.01855 & 0.96892 \\
        \hline
        
        0.7 & 0.3 & 78.77452 & 129.85588 & 0.96586 \\
        \hline

\end{tabular}
}
\end{table}

\begin{table}
\centering
\caption{Evaluation of the proposed ANN model when trained with wind speed and wind direction data only}\label{tab11}
{
\centering
\begin{tabular}{ |p{2cm}|p{2cm}|p{2cm}|p{2cm}|p{2cm}|}
\hline
Training set size &Testing set size &MAE  &RMSE &R Squared\\
\hline
        0.95 & 0.05 & 73.34284 & 136.72901 & 0.96208 \\
        \hline
         
        0.9 & 0.1 & 70.68888 & 127.45645 & 0.96708 \\
        \hline
        
        0.85 & 0.15 & 76.62601 & 125.77426 & 0.96802 \\
        \hline
        
        0.8 & 0.2 & 67.48397 & 116.63269 & 0.97252 \\
        \hline
        
        0.75 & 0.25 & 72.66855 & 120.72298 & 0.97056 \\
        \hline
        
        0.7 & 0.3 & 72.15834 & 121.72978 & 0.96999 \\
        \hline

\end{tabular}
}
\end{table}


\begin{table}
\centering
\caption{Evaluation of the proposed ANN model when trained with wind speed and temperature data only}\label{tab12}
{
\centering
\begin{tabular}{ |p{2cm}|p{2cm}|p{2cm}|p{2cm}|p{2cm}|}
\hline
Training set size &Testing set size &MAE  &RMSE &R Squared\\
\hline
        0.95 & 0.05 & 78.71528 & 146.35033 & 0.95655 \\
        \hline
         
        0.9 & 0.1 & 74.71462 & 131.83461 & 0.96478 \\
        \hline
        
        0.85 & 0.15 & 76.62601 & 125.77426 & 0.96802 \\
        \hline
        
        0.8 & 0.2 & 77.65764 & 129.05494 & 0.96635 \\
        \hline
        
        0.75 & 0.25 & 78.46776 & 127.59082 & 0.96711 \\
        \hline
        
        0.7 & 0.3 & 80.48764 & 130.00208 & 0.96578 \\
        \hline

\end{tabular}
}
\end{table}

\begin{table}
\centering
\caption{Evaluation of the proposed ANN model when trained with wind speed, wind direction, and temperature data}\label{tab13}
{
\centering
\begin{tabular}{ |p{2cm}|p{2cm}|p{2cm}|p{2cm}|p{2cm}|}
\hline
Training set size &Testing set size &MAE  &RMSE &R Squared\\
\hline
0.95 &0.05 &78.84298 &140.94143 &0.95971\\
\hline
0.90 &0.10 &76.75106 &129.84502 &0.96584\\
\hline
0.85 &0.15 &74.36835 &119.89676 &0.97094\\
\hline
0.80 &0.20 &81.07063 &127.50763 &0.96715\\
\hline
0.75 &0.25 &78.28338 &124.01855 &0.96893\\
\hline
0.70 &0.30 &88.00531 &133.50379 &0.96391\\
\hline

\end{tabular}
}
\end{table}
The first thing one can notice from the above tables namely TABLE~\ref{tab10}, TABLE~\ref{tab11}, TABLE~\ref{tab12}, TABLE~\ref{tab13} is the distinct drop in the MAE and RMSE values as well as the high values of the R squared values in comparison to the evaluation tables corresponding to the proposed linear and polynomial regression models. Clearly, the proposed ANN model was capable of eliminating more errors than the parametric models. Additionally, the highest R squared value was obtained when the training data was 85\% of the given data under each condition. The exception to this was in the case of training the proposed ANN model with wind speed and wind direction data only; here, the overall highest R squared value of 0.97252 and overall lowest MAE value of 67.48397 were obtained when the training data was 80\% of the given data taking only wind speed and wind direction parameters into consideration. This means that the predicted values and the actual values were the closest when the proposed ANN model was used for the prediction task and was trained with 80\% of the given data taking only wind speed and wind direction parameters into consideration. From this comparative analysis of the evaluation tables, power curve fitting graphs, and plots of predicted output power versus actual output power obtained for each of the proposed models, it is clear that the proposed ANN model provides a more accurate prediction of wind power production than the proposed linear and polynomial regression models and is, thus, more reliable and suitable for the prediction task for the particular sub-station taken into consideration.

\section{Conclusion}
With the prominence of wind power in the global energy market, it is clear that accurate prediction models are crucial. This is becoming a conspicuous point of research as an important component in operating power systems to maintain their reliability. This paper has presented an outline of the different models that have been attempted for power prediction to showcase the diversity in approaches, and proposed parametric and non-parametric models for performing the same task. While there are various methods to approach the prediction task, the performance of each model depends upon the particular data set with which it should operate. Further, it must be noted that short-term models such as the ones explored in this paper cannot be directly applied to another site. They must be modified by a domain expert in order to take site-specific influencing factors into consideration. By utilizing modelling methods based on the concept of the power curve, the prediction of wind power production has been made convenient. For the data set provided, the proposed ANN model was best-suited to predict wind power production, as it gives the best R squared values and the least RMSE and MAE values. To our knowledge, there are several institutions associated with wind energy production and utilization which use polynomial regression - a model that performed second-best with the given data, as can be seen in this article. This is likely because ANNs take a long time to process larger and larger data sets,whereas polynomial models are simpler to debug.
\section{Declaration of Competing Interest}
The authors declare that they have no known competing financial interests or personal relationships that could have appeared to influence the work reported in this paper.

\end{document}